%% file: 0_asr_aware_diarization.tex
\documentclass[9pt]{extarticle}
\usepackage{spconf,amsmath,graphicx,xcolor, multirow}
\usepackage{booktabs}
\usepackage{hyperref}
\usepackage{graphicx}
\usepackage{comment}
\usepackage{xcolor}
\usepackage{graphicx}
\usepackage{subcaption}
\usepackage[normalem]{ulem}
\usepackage{amsfonts}

\title{ASR-aware end-to-end neural diarization}
\name{Aparna Khare, Eunjung Han, Yuguang Yang, Andreas Stolcke}
\address{
	{Amazon Alexa AI, Sunnyvale, CA}\\
	{\small\texttt{apkhare,cehan,yuguay,stolcke@amazon.com}}}
\begin{document}
	%
	\maketitle
	\begin{abstract}	
		We present a Conformer-based end-to-end neural diarization (EEND) model that uses both acoustic input and features derived from an automatic speech recognition (ASR) model. Two categories of features are explored: features derived directly from ASR output
		(phones, position-in-word and word boundaries) and features derived from a lexical speaker change detection model, trained by fine-tuning a pretrained BERT model on the ASR output. Three modifications to the Conformer-based EEND architecture are proposed to incorporate the features. First, ASR  features are concatenated with acoustic features. Second, we propose a new attention mechanism called contextualized self-attention that utilizes ASR features to build robust speaker representations. Finally, multi-task learning is used to train the model to minimize classification loss for the ASR features along with diarization loss. Experiments on the two-speaker English conversations of Switchboard+SRE data sets show that multi-task learning with position-in-word information is the most effective way of utilizing ASR features, reducing the diarization error rate (DER) by 20\% relative to the baseline.
	\end{abstract}

    \input{1_intro}

	\input{2_conformer_based_diarization}

    \input{data_figure1}

	\input{3_asr_features_for_diarization}
	\input{4_methods_for_incorporating_asr_features}

\input{data_table1}

	\input{5_data_and_experiments}
	\input{6_results_and_discussion}

	\input{7_conclusion}
	

	\bibliographystyle{IEEEbib}
	\bibliography{refs}
	
\end{document}

%% file: 1_intro.tex
\section{Introduction}
\label{sec:intro}
Speaker diarization addresses the question 'who spoke when' in multi-speaker conversations. It is useful for applications such as speaker-attributed ASR \cite{kanda2019acoustic,boeddeker2018front}, and analyzing and indexing recorded meetings \cite{vijayasenan2009information}. Conventional diarization systems are based on clustering speaker embeddings from short speech segments. The typical design extracts speaker embeddings from short speaker-homogeneous segments in the first stage, followed by clustering, to identify segments from the same speakers \cite{sun2021content}. More recently, deep learning models like EDA-EEND \cite{horiguchi20_interspeech} and SA-EEND \cite{fujita2019end} allow training an end-to-end diarization model with state-of-the-art performance. Most diarization systems use speech \cite{zhang2019fully} or audio-visual inputs \cite{gebru2017audio}. Intuitively, the lexical content in the audio could provide valuable information to aid the task. Backchannel words and pause fillers like `uh-huh' or 'mhm' indicate the other speaker is likely to keep talking \cite{sacks1978simplest}. Sun et al.\ verify this intuition in \cite{sun2021content} by extracting content-aware speaker representations for a clustering-based diarization system. Park et al.\ use lexical information to estimate speaker turn probabilities to improve performance \cite{park2019speaker,park18b_interspeech}. Anidjar et al. use lexical information using word2vec embeddings to build a speaker change detection model in \cite{anidjar21_interspeech}. Still, prior work has not explored the use of ASR features in end-to-end neural diarization (EEND) models. Additionally, these prior studies are largely limited to simple concatenation of ASR-derived and acoustic features for diarization.
	
	In this paper, we explore if ASR features can improve performance of Conformer-based EEND systems. Our focus is on two-speaker conversations in English, as we use a hybrid English ASR model. Our main contributions are: (1) The use of time-aligned phone, position-in-word information, and word boundaries as additional features for diarization. (2) Use of a pretrained BERT model fine-tuned for lexical speaker change detection as a feature extractor. (3) Comparing three different methods for leveraging ASR features:
	concatenation with acoustic features, providing them as context in the attention module of a Conformer-based EEND model (contextualized self-attention), and multi-task learning.

%% file: 2_conformer_based_diarization.tex
\section{Conformer-based diarization}
\label{sec:baseline}
Our baseline model is a Conformer-based EEND model described in \cite{liu21j_interspeech}. The acoustic input to the model $X_{\text{Acoustic}}$ is subsampled by a factor of 4 with a 2-D convolutional layer. A Conformer encoder with multiple layers is used to obtain frame-level embeddings $E$. Each Conformer layer has a multi-headed self-attention layer, a convolution layer and a linear layer, with residual connections to all the layers. Frame-wise posteriors of speaker activity for all speakers are estimated from $E$ using a linear layer of output dimension equal to the number of speakers $S$, followed by a sigmoid activation. This allows the model to estimate overlapped speech from multiple speakers. To deal with ambiguity in which class represents which speaker,  a training scheme that considers all permutations of the reference speaker labels is used. Experiments were limited to two-speaker conversations, i.e. $S=2$. The model is trained to minimize diarization loss $\mathcal{L}_{DER}$, which is computed as sum of the binary cross entropy loss over all speakers $S$.

%% file: data_figure1.tex
\begin{figure*}[htb]
\begin{subfigure}{0.33\textwidth}
  \centering
  \includegraphics[width=0.87\linewidth]{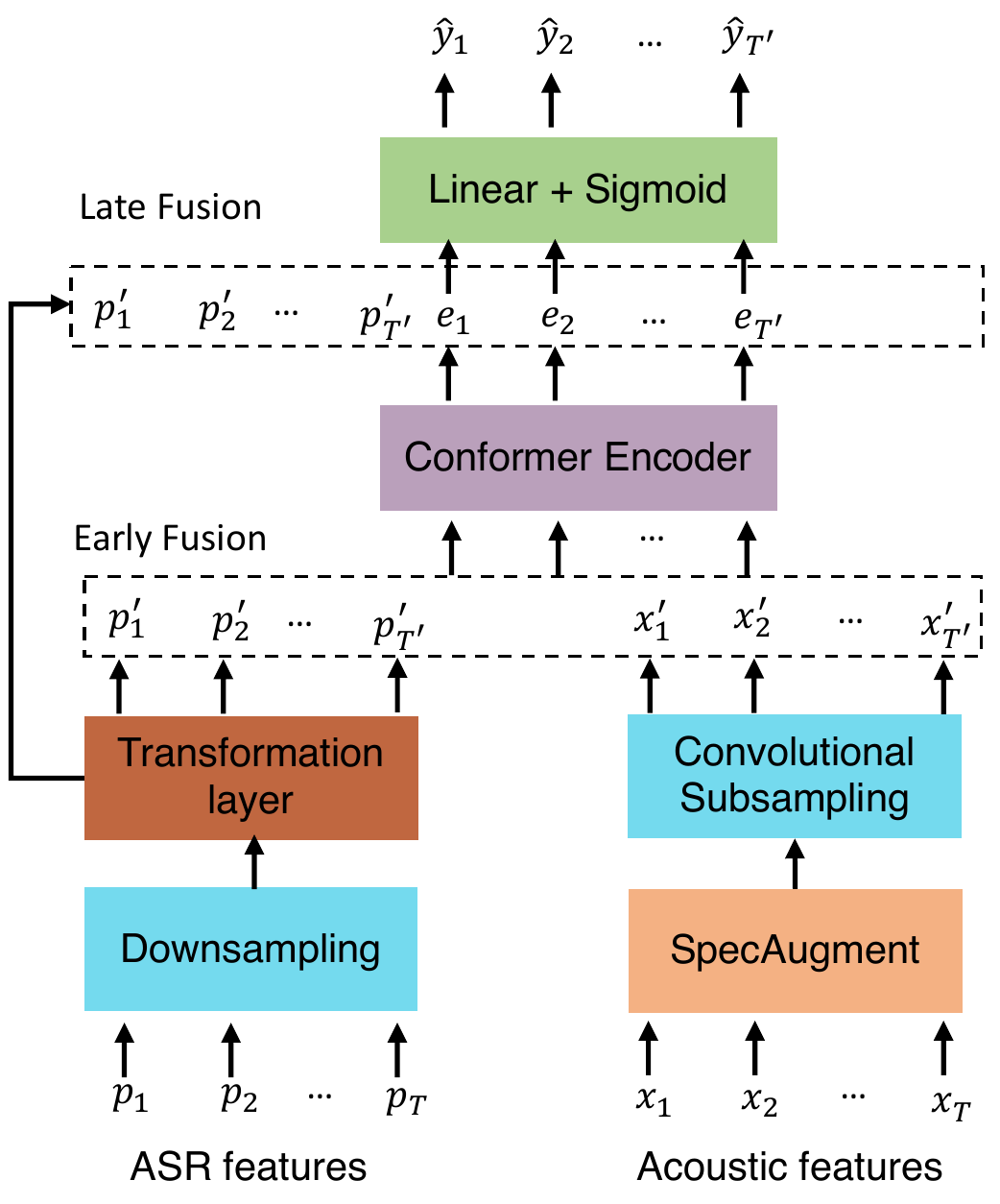}
  \caption{Concatenation}
  \label{fig:sfig2}
\end{subfigure}
\begin{subfigure}{0.33\textwidth}
  \centering
  \includegraphics[width=0.96\linewidth]{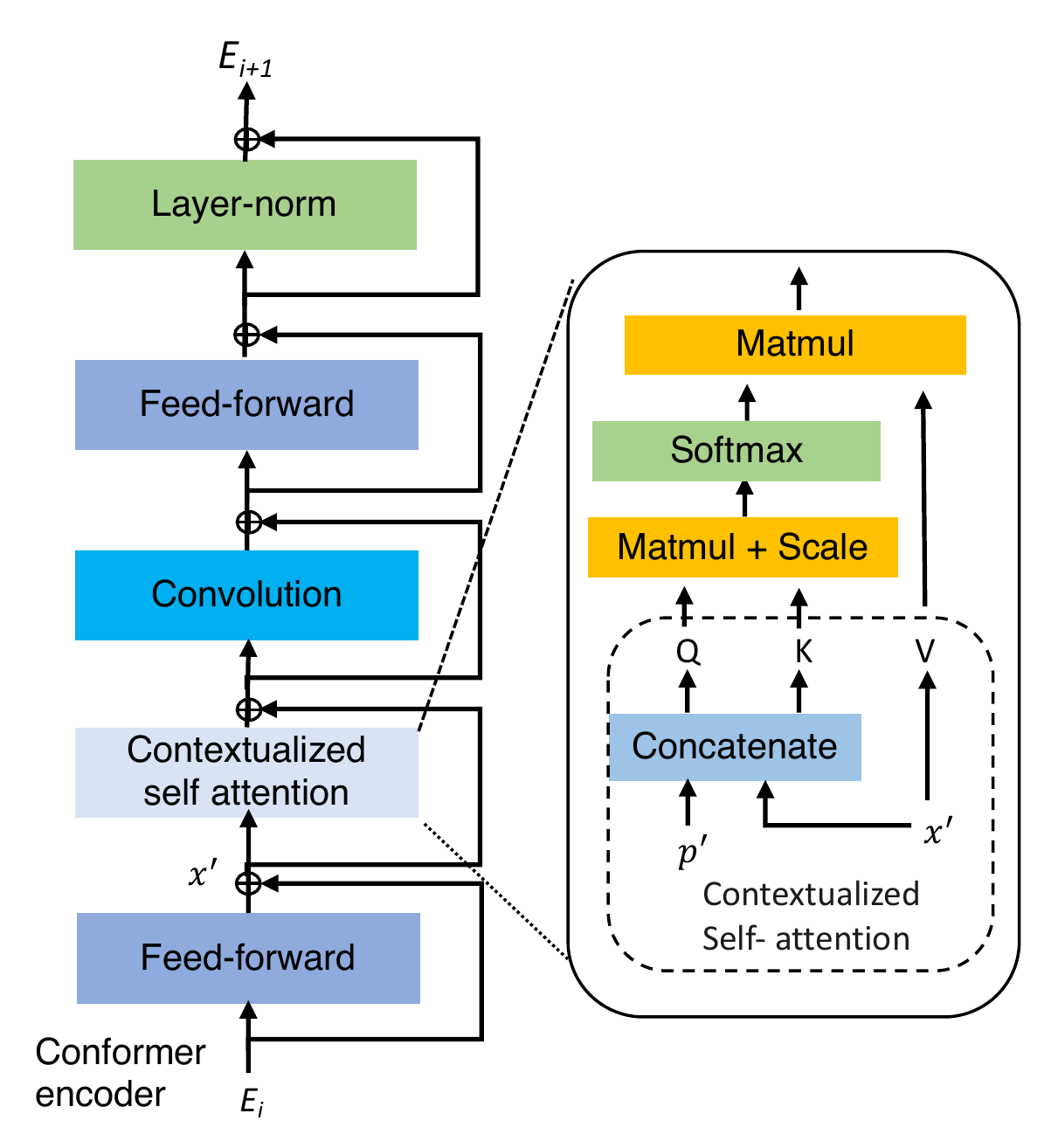}
  \caption{Contextualized self-attention}
  \label{fig:sfig1}
\end{subfigure}%
\begin{subfigure}{0.33\textwidth}
  \centering
  \includegraphics[width=0.84\linewidth]{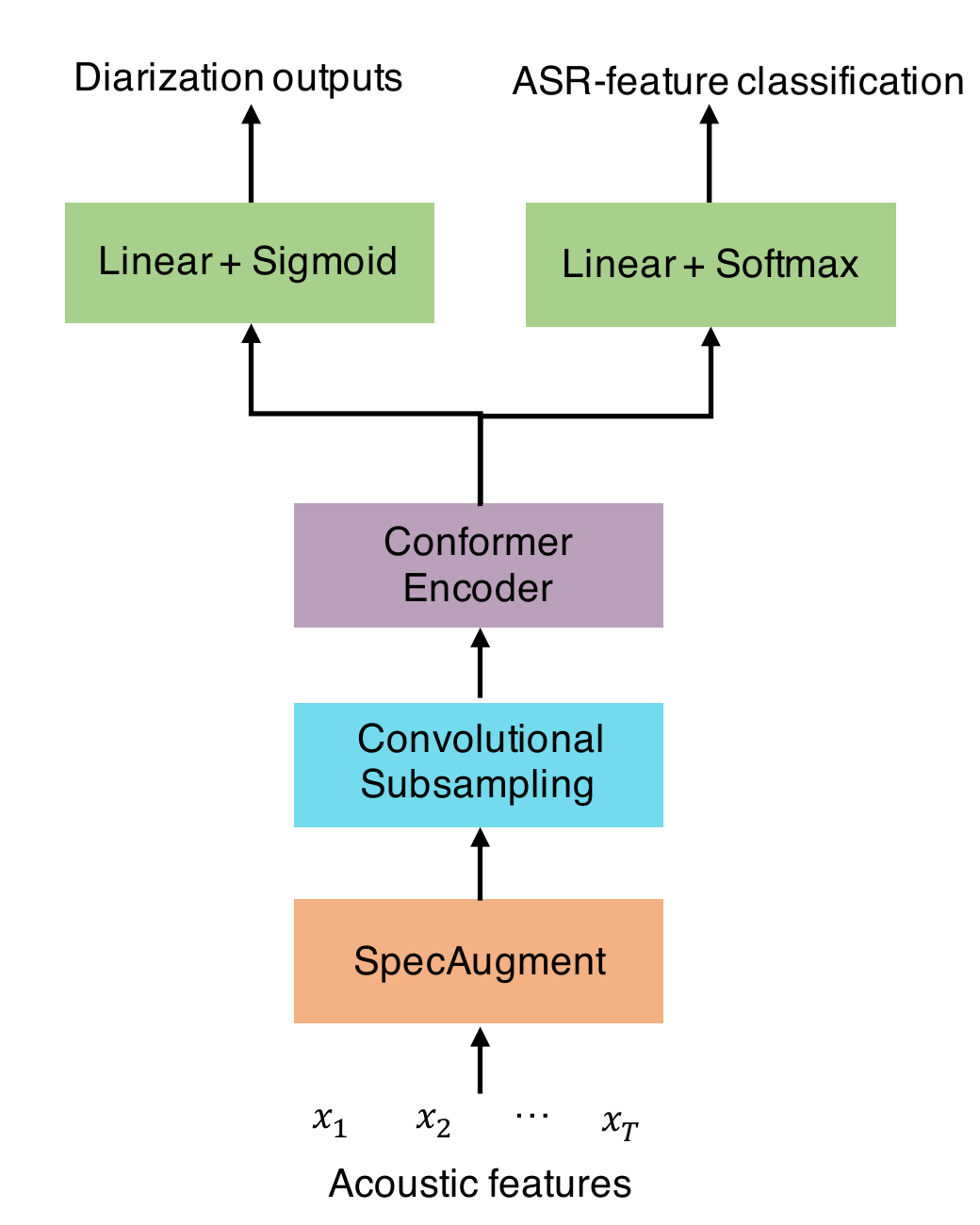}
  \caption{Multi-task training with ASR features}
  \label{fig:sfig3}
\end{subfigure}
\vspace{-2mm}
\caption{Three methods for incorporating ASR features into diarization}
\label{fig:method}
\vspace{-4mm}
\end{figure*}

%% file: 3_asr_features_for_diarization.tex
\section{ASR features for diarization}
\label{sec:features}

\subsection{Features from ASR model}

Three frame-level features are identified for diarization directly from the ASR output:

\vspace{0.1cm}
\noindent \textbf{Phones}: \cite{sun2021content} shows that phone labels can improve diarization models. Phone information allows aggregation of features from phonetically similar frames and helps build robust speaker representations by taking content variability into account.

\vspace{0.1cm}
\noindent \textbf{Position-in-word}: Hybrid ASR models typically use word position dependent phones \cite{povey2011kaldi}, as the position of the phone in the word can alter its pronunciation. The position-in-word features used in our experiments are  represented by 5 classes; silence, singleton, begin, internal and end, as defined in the Kaldi toolkit. Position-in-word for each frame informs the model if the frame is at the beginning, in the middle or at the end of the word. If the frame is at the beginning or in the middle of a word, the model decision is less likely to change from the previous frame. In addition, the silence phone labels help with the prediction of non-speech (no speaker activity).

\begin{figure}
\includegraphics[scale=0.18]{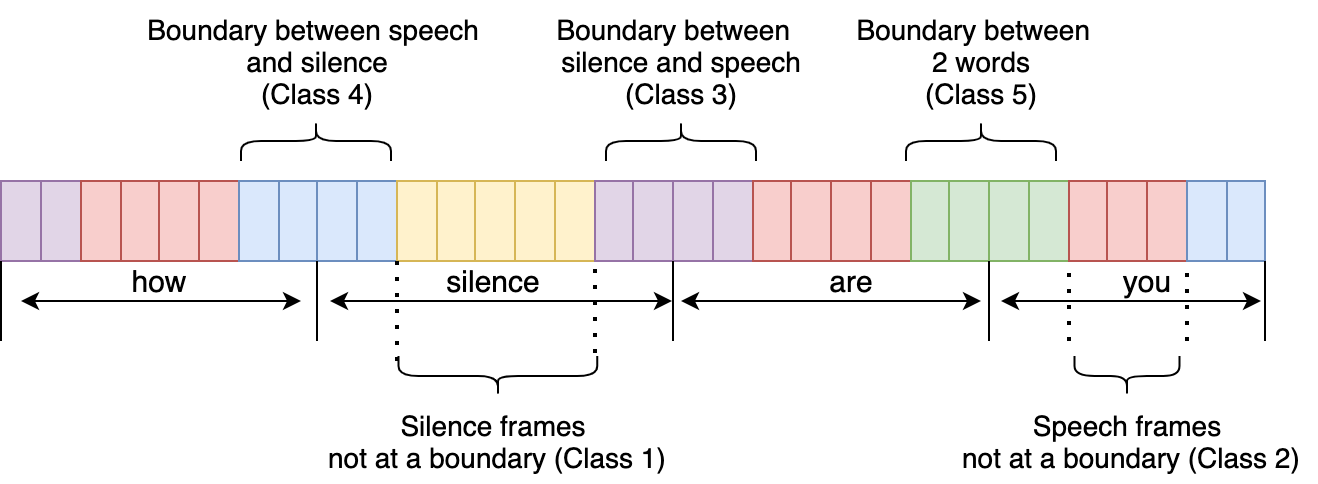}
\caption{Example of word boundary feature extraction.}
\label{fig:wbfeats}
\vspace{-5mm}
\end{figure}

\vspace{0.1cm}
\noindent \textbf{Word boundaries}: A speaker change is more likely to happen at a word boundary, so these features are likely to improve diarization. Hybrid ASR models can predict the phone boundaries  with over 90\% accuracy; we assume a low error rate in boundary detection for words that ASR detects correctly \cite{stolcke2014highly}. We model the word boundary as a 5-dimensional input as shown in Figure~\ref{fig:wbfeats}.
The final feature is a 1-hot vector representing the class assigned to the frame.
The classes are: (1) silence frames not at any boundary, (2) speech frames not at a boundary, (3) boundary between silence frame followed by speech frames, (4) boundary between speech frame followed by silence, and (5) boundary between two words. Two frames before and after a boundary are assigned to the appropriate class, such that a boundary is represented by four frames. We apply downsampling to ASR features to match the acoustic sample rate, described in Section~\ref{ssec:concat}, and a four frame word boundary feature prevents it from missing a boundary frame, as it retains every fourth frame in the utterance.

All features are at the frame-level to match the acoustic input for diarization. We also performed an experiment with binary speech activity input derived from the frame-level phone alignment to understand how much improvement can be attributed to speech activity detection compared to the more granular features.

\subsection{Lexical speaker change features}
Certain words or phrases in a conversation can be associated with phenomena that signal the speaker's intention of taking a turn \cite{xiong2018session}. We aim to learn these features based on the ASR 1-best hypothesis and use them for diarization.
	
Specifically, we formulate speaker change detection as a sequence labeling problem. Let $\mathbf{x}=\left(x_{1}, x_{2} \ldots x_{T}\right)$ be the sequence of $T$ tokens (after tokenization) from the ASR hypothesis. The target sequence for the speaker change detection task is denoted by $\mathbf{y}=\left(y_{1}, y_{2} \ldots y_{T}\right) \in\{0,1\}^{T}$, where  $y_{i}=1$ if token $i$ is the first token following a speaker change, and 0 otherwise. We fine-tune a pre-trained BERT model \cite{sanh2019distilbert} for this task, which outputs the probability of speaker change prior to this token. Details of the model architecture and training are provided in Section \ref{sec:data_exp}. The following features are extracted from this model: 

\vspace{0.1cm} 
\noindent \textbf{Speaker change posteriors (SCP)}: This is a 2-dimensional feature, with the speaker change posteriors obtained from the lexical speaker change model as the first dimension, and the non-speech activity obtained from the ASR word alignments as the second. The posterior feature is not available for time segments without associated recognized words. For those segments, the posterior value is set to 0 and the non-speech activity is set to 1. All frames corresponding to a given word interval are represented by the same feature.

\vspace{0.1cm}
\noindent \textbf{Speaker change embeddings (SCE)}: The output of the BERT encoder before the classification layer is used, as these features may capture semantic information relevant for speaker change detection. 
\vspace{-5mm}

%% file: 4_methods_for_incorporating_asr_features.tex
\section{Methods for incorporating ASR features}
\label{sec:methods}

\subsection{Concatenation}
\label{ssec:concat}
Concatenation of ASR features with the acoustic features has shown improvements in diarization performance in prior work \cite{sun2021content}. 
Given audio features $X$, we first generate \textit{acoustic features} $X_{\text{Acoustic}}$ by applying SpecAugment on $X$. Then, we compute $X'_{\text{Acoustic}}$ by convolutional subsampling of $X_{\text{Acoustic}}$.
We obtain \textit{ASR features} $X_{\text{ASR}}$ (described in Section \ref{sec:features}) from the ASR hypothesis. Based on the feature, we use different functions (\textit{Transformation layer} in Figure~\ref{fig:sfig2}) to obtain the input to the model. For the directly derived ASR features, the transformation layer is an Embedding layer that maps the ASR features from a one-hot input to an embedding. For the 768-dimensional speaker change embeddings, the transformation layer consists of a feed-forward layer of size 64 followed by a ReLU activation before concatenating these features.  
The speaker change posteriors are used directly (identity transformation). We then subsample the ASR features to compute $X'_{\text{ASR}}$ to match the sampling rate of $X'_{\text{Acoustic}}$. We apply two methods for downsampling: simple downsampling and separate convolutional subsampling layers.  Simple downsampling is lossy in nature as we drop $n-1$ out of $n$ frames, where $n$ is the downsampling factor. Convolutional subsampling can address this drawback and can capture local similarities in features.
\begin{equation*}
\begin{split}
X_{\text{Acoustic}} &= \text{SpecAugment}(X) \\
X'_{\text{Acoustic}} &= \text{ConvolutionalSubsampling}(X_{\text{Acoustic}}) \\
X_{\text{ASR}} &= \text{Transformation}(\text{FeatExtractModel}(X)) \\
X'_{\text{ASR}} &= \text{Downsampling}(X_{\text{ASR}}) \\
\end{split}
\end{equation*}
where FeatExtractModel is either the ASR model or the lexical speaker change detection model. Figure~\ref{fig:sfig2} shows the two methods of concatenating the features.

\noindent \textbf{Early fusion}: Providing the ASR features as input to the Conformer encoder would allow the Conformer layers to build context-aware speaker embeddings, allowing for better diarization. For early fusion, we concatenate the subsampled ASR features $X'_{\text{ASR}}$ with the input acoustic features $X'_{\text{Acoustic}}$.
\begin{equation*}
\begin{split}
X_{\text{Concat}} &= \text{Concatenate}(X'_{\text{Acoustic}}, X'_{\text{ASR}}) \\
E &= \text{ConformerEncoder}(X_{\text{Concat}})  \\
Z &= \sigma(\text{Linear}(E))  
\end{split}
\end{equation*}
where $E$ is $D$-dimensional diarization embeddings and $Z$ represents frame-wise posteriors of the joint speech activities for $S$ speakers.

\noindent \textbf{Late fusion}: High level features like speaker change posteriors could be utilized more efficiently by the classification layer to determine speaker change points. For late fusion, we concatenate the subsampled ASR features $X'_{\text{ASR}}$ with the output of the Conformer $E$.
\begin{equation*}
\begin{split}
E &= \text{ConformerEncoder}(X'_{\text{Acoustic}})  \\
E_{\text{concat}} &=  \text{Concatenate}(E, X'_{\text{ASR}}) \\
Z &= \sigma(\text{Linear}(E_{\text{concat}}))  
\end{split}
\end{equation*}
\subsection{Contextualized self-attention (CSA)}	
ASR features provide context to the model for making better decisions. However, adding the low-level features, such as phone labels, directly to the Conformer layer may not allow the model to use these features just as context and can add noise. Therefore, we introduce contextualized self-attention.

Our algorithm is illustrated in Figure~\ref{fig:method}(b). Given the Conformer encoder with $n$ layers, let $E^{i}$ denote the output of the $i$th layer. The subsampled ASR features,  $X'_{\text{ASR}}$, are concatenated with the query $Q^i$ and key $K^i$ at each layer of the Conformer model; the value $V^i$ is only constructed using $E_{i-1}$. 
This architecture allows a frame to find similar frames in the sequence based on both acoustic and contextual similarity, but the attended features passed to the next layers only contain acoustic information as the output is a scaled version of the value $V^i$.  The final diarization embedding is $E = E^n$.
\begin{equation*}
\begin{split}
V^{i} &= E^{i-1} \mbox{ for } 1 \le i \le n \ (E^{0} = X'_{\text{Acoustic}}), \\
K^{i} &= Q^{i} = \text{Concatenate}(X'_{\text{ASR}}, E^{i-1}) \mbox{ for } 1 \le i \le n, \\
E^i &= \text{ConformerEncoderLayer}(Q^i,K^i,V^i) \mbox{ for } 1 \le i \le n,  \\
\end{split}
\end{equation*}

\vspace{-0.5cm}

\subsection{Multi-task learning}

Multi-task learning with related tasks has been found beneficial in tasks like keyword spotting \cite{panchapagesan2016multi} and speech recognition \cite{paraskevopoulos2020multimodal}. Multi-task learning with phonetic information has been successfully applied to the speaker verification task as well \cite{chen2020channel,liu2019introducing}. Our proposed multi-task method is shown in Figure~\ref{fig:sfig3}. The ASR features are used as labels for auxiliary tasks that the model is trained on. The features learned by the Conformer layers in this framework are ASR-aware, and implicitly utilized for diarization. As an added advantage, this method eliminates the need for an ASR system during inference. The output of the $i^{th}$ Conformer layer is passed to a 2-layer linear network with ReLU activation to compute the logits for the  ASR feature class. The additional loss $\mathcal{L}_{\text{AUX}}$ is computed as the cross-entropy loss for the ASR-feature targets. The final loss $\mathcal{L}$ is computed as 
\vspace{-1mm}
\begin{equation*}
	\mathcal{L} =\mathcal{L}_{\text{DER}} + \alpha \mathcal{L}_{\text{AUX}}
\end{equation*}
The Conformer layer index $i$ and the hyper-parameter $\alpha$ are tuned on a held out development set.
\vspace{-2mm}

%% file: data_table1.tex
\begin{table*}[t]
\caption{DER (\%) on SWB + SRE test set. Relative DER improvement (\%) is shown in brackets. Best DER for each feature is shown in bold. CSA is contextualized self-attention, SCP is speaker change posteriors, and SCE is speaker change embeddings from the speaker change model. Baseline DER with acoustic features only is 3.19\%.}
\vspace{-3mm}
\label{table:ders}
\centering
\begin{tabular}{l|l|c@{\hskip 0.5mm}r c@{\hskip -2mm}r c@{\hskip -10mm}r |c@{\hskip 0.5mm}r c@{\hskip 0.5mm}r}
	\toprule
	Method & Downsampling & \multicolumn{2}{c}{Phones} & \multicolumn{2}{c}{Position-in-word} & \multicolumn{2}{c|}{Word boundaries} & \multicolumn{2}{c}{SCE} & \multicolumn{2}{c}{SCP} \\
    \midrule
	Early fusion &  Simple downsampling       
	& 3.28 & (-2.82\%) & 3.09 &  (3.13\%) & 3.38 &  (-5.96\%) 
	& 3.22 &  (-0.94\%) & 3.11 &  (2.51\%)  \\
	Early fusion & Conv-subsampling 
	& 2.90 &  (9.09\%) & 2.87 &  (10.03\%) & 2.97 &  (6.90\%)  
	& 2.94 &  (7.84\%) & \multicolumn{2}{c}{-}\\
 	Late fusion & Simple downsampling 
 	& 2.89&   (9.40\%) & 3.14 &  (1.57\%) & 3.03 &  (5.02\%)  
 	& 3.07 &  (3.76\%) & \textbf{2.88} &  \textbf{(9.72\%)}	 \\
 	Late fusion & Conv-subsampling 
 	& 2.80&   (12.23\%) & 2.89 &  (9.40\%) & 3.11 &  (2.51\%)
 	& \textbf{2.89} &  \textbf{(9.40\%)} &  \multicolumn{2}{c}{-}	 \\
	\midrule
	CSA & Simple downsampling     
	& 3.09 &  (3.13\%)  & 2.86 &  (10.34\%) & 2.97 &  (6.90\%) 
	& 3.17 &  (0.63\%) & 3.26 &  (-2.19\%)  \\
	CSA & Conv-subsampling 
	& 2.99 &  (6.27\%) & 2.93 &  (8.15\%)  & 3.03 &  (5.02\%) 
	& 3.02 &  (5.33\%) &  \multicolumn{2}{c}{-} \\
	\midrule
	Multi-task  & -
	&  \textbf{2.63} &  \textbf{(17.55\%)} & \textbf{2.54} & \textbf{(20.38\%)} &  \textbf{2.60} & \textbf{(18.50\%)}
	&  \multicolumn{2}{c}{-} &  \multicolumn{2}{c}{-}\\
	\bottomrule
\end{tabular}
\vspace{-3mm}
\end{table*}

%% file: 5_data_and_experiments.tex
\section{Data and Experiments}
\label{sec:data_exp}

\vspace{0.1cm} 
\noindent \textbf{Data}: 
All experiments are performed on the two-speaker subset from two-channel Switchboard-2 (Phase I, II, III), Switchboard Cellular (Part 1, 2) (SWB), and the 2004-2008 NIST Speaker Recognition Evaluation (SRE) data sets. All recordings are telephone speech data sampled at 8 kHz and mapped to a single channel. Since these data sets do not contain time annotations, we extract speech segments using speech  activity  detection  (SAD)  on  the  basis  of  a  time-delay neural network and statistics pooling from a Kaldi recipe to obtain the diarization labels. The language information for selecting English data is obtained from the sphere file headers. The final data set consists of 5,416 speakers. It is partitioned into train, dev and test partitions with no speaker overlap between the 3 sets. The training set has 19,589 recordings with 3,861 speakers (1687 hours), the dev set has 955 recordings with 876 speakers (80 hours), and the test set has 1,501 recordings with 679 speakers (126 hours). 

\vspace{0.1cm} 
\noindent \textbf{Conformer-based EEND}: 
The acoustic features are 80-dimensional log filter-bank energies (LFBE) extracted with a frame size of 25\,ms and a frame shift of 10\,ms. SpecAugment \cite{park2019specaugment} is applied to the LFBE features with two frequency masks with a maximum mask size of 2, and two time masks with a maximum mask size of 1200. 
Two 2-dimensional (2D) depth-wise separable convolutional subsampling layers downsample the LFBE features by four (corresponding to 40\,ms), with a kernel size of $\{(3, 3),(7, 7)\}$ and stride of $\{(2, 2),(2, 2)\}$. For convolution sub-sampling of ASR features, similar subsampling with separate layers is applied. The parameters for two layers of the 2D convolution are: $\{(3, 3), (7, 3)\}$ for kernels and $\{(2, 2), (2, 2)\}$ for strides.
Four Conformer layers of size 256 with 4 attention heads are used in the encoder.  No positional encodings are used based on the finding in \cite{liu21j_interspeech}. All models are trained with a batch size of 192 with the Adam optimizer, with the learning rate scheme described in \cite{vaswani2017attention} for 200 epochs. The final model used for evaluation is obtained by averaging model parameters from epochs 191 to 200. Diarization error rate (DER) is used to evaluate model performance. As is standard, a tolerance of 250 ms when comparing hypothesized to reference speaker boundaries was used.

\vspace{0.1cm} 
\noindent \textbf{ASR model}: 
A pre-trained%
\footnote{\href{https://kaldi-asr.org/models/1/0001_aspire_chain_model_with_hclg.tar.bz2}{https://kaldi-asr.org/models/1/0001\_aspire\_chain\_model\_with\_hclg.tar.bz2}} ASR model based on a time-delay neural network (TDNN) trained on multi-condition Fisher English data is used to extract ASR features. This model is trained on conversational English data and suits our requirements. The time aligned word and phone sequences are obtained from the 1-best ASR hypothesis for both training and evaluation. The ASR model uses a 54-dimensional phone set with word position dependent phones. Frame-level phone labels are obtained by collapsing all the word position dependent representations of a specific phone into a single phone. The position-in-word labels are obtained from the phone suffixes in the alignment files. For concatenation and contextualized self-attention experiments, a 16-dimensional embedding layer is used to learn feature representations. For the multi-task experiments, the linear layer used for classification is of dimension 256. 

\vspace{0.1cm} 
\noindent \textbf{Lexical speaker change detection model}: 
The lexical speaker change detection model is obtained by fine-tuning a pretrained DistilBERT\footnote{\href{https://huggingface.co/distilbert-base-uncased}{https://huggingface.co/distilbert-base-uncased}} model\cite{sanh2019distilbert}. 
We tokenize the ASR transcriptions with the DistilBERT uncased tokenizer. The lexical speaker change model output is at the sub-word level, but the ASR time aligned output is at word-level. To get sub-word time alignments, the time interval for a word is split into $n$ equal intervals, each corresponding to the $n$ sub-word units in that word. The sub-word level time alignments are then used to align speaker change features with audio frames. We use two special tokens to represent the \emph{noise} and \emph{laugh} tokens in the transcriptions. The first token in a given speaker turn is labeled as \textit{speaker-change tokens}, meaning that they signal the change of a speaker's turn; other tokens are labeled as \textit{no-speaker-change tokens}. The network is trained using weighted cross-entropy loss with class weights 0.065 and 0.935 for speaker-change and no-speaker-change classes respectively, to compensate for the imbalanced class priors. The model accuracy on the development set is 77\%.
The model is trained on the training partition of the SWB+SRE data for 20000 steps, with a batch size of 40. Each training sequence consists of 20 consecutive tokens randomly sampled from a transcript. During inference. we use a sliding window of 20 input tokens, with an overlap of 10. For each window, we only use the speaker-change features for the 10 tokens in the middle to ensure that each token has past and future context of at least 5 tokens.

%% file: 6_results_and_discussion.tex
\section{Results and discussion}
\label{sec:results}

Table \ref{table:ders} shows the results of our experiments. Multi-task training yields the best results out of the three methods for all ASR features. Position-in-word features gives the best results, with a 20.38\% relative improvement in DER. The tuned hyper parameters for the multi-task architecture are as follows. The best Conformer layer $i$ used for computing the auxiliary loss was the final layer. The optimal $\alpha$ for weighting the ASR loss for position-in-word features is 0.2, and 0.6 for the phone and word boundaries. We do not perform multi-task learning with speaker change features as the speaker change information is already subsumed in the diarization loss function.  

For both early and late fusion, convolutional subsampling outperforms simple downsampling, with the exception of late fusion with word boundaries. For contextualized self-attention, simple downsampling is better for position-in-word and word boundaries. We posit that convolution may add some noise to the categorical features, leading to noisy attention weights. In late fusion, phone features outperform other features, improving DER by 12.23\% relative. For position-in-word features, contextual self-attention outperforms fusion methods, improving the DER by 10.34\% relative, validating our hypothesis that adding this feature only as context helps build robust embeddings for diarization. As with multi-task training, position-in-word features are the most informative except with late fusion. Contextualized self-attention with simple downsampling using SAD features alone results in a DER of 3.11\% (2.51\% relative improvement, results not in the table). It shows that improvement with other features is based on information more granular than speech activity.

Next, we look at improvements from the lexical speaker change detection model. Late fusion with speaker change posteriors gives the best results, with a relative improvement of 9.72\%, as this feature directly informs the classification layer whether there is a speaker change at the current frame or not. Late fusion with speaker change embeddings also improves the performance by 9.40\% relative.
 
 We did an additional experiment to combine late fusion with speaker change posteriors and multi-task learning with position-in-word features to check whether the techniques are complementary. This experiment did not yield additional improvements over multi-task learning alone.
\vspace{-2mm}

%% file: 7_conclusion.tex
	\section{Conclusion}
	\label{sec:conclusion}
	We have shown that position-in-word, phones, word boundary features and features from a lexical speaker change detection model can improve the DER in a Conformer-based EEND model, with position-in-word being the most useful ASR feature. We show that multi-task training is the most effective way to utilize these features, with a DER reduction of up to 20\% relative. A new contextualized self-attention mechanism is introduced, reducing DER rate with position-in-word features by 10\% relative, outperforming simple concatenation.
	
	In future work, we hope to explore similar techniques with a system like \cite{han2020bw} that does not require prior knowledge of the number of speakers and allows inference with low latency. Note that state-of-the-art ASR systems like RNN-Transducer \cite{he2019streaming} cannot be used to extract phone and word-position information, as they output sub-word units and not phones. Our work is further evidence that future work should aim to treat ASR and diarization as a joint task.